This is the accepted version of an article that has been published in 2023 8th International Conference on Control and Robotics Engineering (ICCRE). Changes were made to this version by the publisher prior to publication. The final version of record is available at https://doi.org/10.1109/ICCRE57112.2023.10155581

# Task segmentation based on transition state clustering for surgical robot assistance

Yutaro Yamada
Department of Micro-Nano Mechanical Science and Engineering
Nagoya University
Nagoya, Japan
yamada@robo.mein.nagoya-u.ac.jp

Jacinto Colan
Department of Micro-Nano Mechanical Science and Engineering
Nagoya University
Nagoya, Japan
colan@robo.mein.nagoya-u.ac.jp

Ana Davila
Institutes of Innovation for Future Society
Nagoya University
Nagoya, Japan
davila.ana@robo.mein.nagoya-u.ac.jp

Yasuhisa Hasegawa
Department of Micro-Nano Mechanical Science and Engineering
Nagoya University
Nagoya, Japan
hasegawa@mein.nagoya-u.ac.jp

***Abstract*— Understanding surgical tasks represents an important challenge for autonomy in surgical robotic systems. To achieve this, we propose an online task segmentation framework that uses hierarchical transition state clustering to activate predefined robot assistance. Our approach involves performing a first clustering on visual features and a subsequent clustering on robot kinematic features for each visual cluster. This enables to capture relevant task transition information on each modality independently. The approach is implemented for a pick-and-place task commonly found in surgical training. The validation of the transition segmentation showed high accuracy and fast computation time. We have integrated the transition recognition module with predefined robot-assisted tool positioning. The complete framework has shown benefits in reducing task completion time and cognitive workload.**

***Keywords—robot assisted surgery, minimally invasive surgery, surgical task segmentation, hierarchical clustering.***

## I. Introduction

Robotic systems are increasingly being incorporated into minimally invasive surgical (MIS) procedures, leading to fewer injuries by performing operations through small incisions. Robotic surgery overcomes some limitations found in MIS [1], providing enhanced image processing for better visibility of tissues and organs, and increasing intuitiveness and force feedback for a precise tool control to compensate for human error.

Current surgical robotic systems are teleoperated by a trained surgeon using a remote console or interface [2]. However, ensuring the safety of patients during surgery is a critical aspect that relies on many factors, and surgical errors can occur due to human shortcoming, such as information loss, poor decision-making, lack of dexterity, fatigue, or lack of attention. Automated image and data analysis robotic systems can aid human operators in reducing errors [3]. To ensure the robot can perform tasks safely and reliably, it must possess the necessary skills and precision to execute the required movements.

One approach to facilitate robotic planning is to segment surgical tasks by breaking down complex tasks into motion primitives and identifying the start and end times of each primitive [4], [5]. This segmentation of the kinetic trajectory can also provide a more quantitative description of technical skill development than relying on global performance indicators such as completion time or total path length [6], [7].

However, segmenting surgical tasks presents several challenges due to the subjectivity and smooth transition of segments, making it challenging to generate consistent segments as data grows over time. Therefore, researchers are developing ways to automate annotation processes using video and tool kinematics, individually or in combination [8], [9], [10].

Current proposals for learning segmentation criteria with minimal supervision (i.e., no labeling) rely on state representation, which is particularly difficult for visual perception pipelines [11]. However, in a well-defined surgical training procedure, the same sequence of segments is typically present, allowing common sets of transition states to occur at similar times and conditions. This consistent and repeated structure can be used to deduce global segmentation criteria for the surgical task.

In this paper, we introduce a transition-based segmentation recognition approach that aims to facilitate a specific robotic assistance in surgical training tasks. By clustering the kinematic and visual information, we can identify segments that share common characteristics and reduce the need for explicit labeling of the data. To validate our approach, we implement it on a robotic system designed for pick-and-place surgical training tasks. We evaluate the effectiveness of our approach in improving the accuracy and efficiency of the robotic system's performance.

## II. Related Works

In recent years, there have been attempts to automatically segment surgical tasks. These efforts have focused mainly on identifying transitions between segments. One such approach is Transition State Clustering (TSC) [12], an unsupervised segmentation algorithm that employs a hierarchical Dirichlet process Gaussian mixture model to analyze surgical kinematic data. TSC aims to identify spurious segmentation endpoints generated by inconsistent motion, under the assumption that the action order is partially consistent throughout the surgical procedure. Using a dataset of kinematic recordings with manually annotated visual features of surgical needle passing and suturing, TSC was able to recognize 83% of needle passing transitions and 73% of suturing transitions annotated by human experts. The integration of kinematic signals with visual features extracted using a pretrained CNN further improved segmentation performance [13]. Another proposed approach is the Dense Convolutional Encoder-Decoder Network (DCED-Net), which is designed to extract dimensionally reduced visual features for surgical task segmentation [14]. Employing an iterative algorithm to merge adjacent segments that were deemed similar, as part of a refinement process aimed at eliminating spurious transitions.

The bottom-up clustering approach leverages the sequential nature of motion signals to merge neighboring segments based on different stopping and merging criteria. Soft-USG [15], automatically segments surgical gestures based on their gradual transitions, utilizing fuzzy membership scores. Soft-USG utilizes kinematic trajectory data, which reduces computation time and makes it more practical for real-time segmentation in robotic surgery. Unlike other methods that combine similar data samples or segments, Tsai et. al, introduced an approach that searches for optimal segmentation criteria. It uses spatio-temporal and variance properties derived from kinematic data to identify potential segmentation points, followed by a clustering-based refinement. However, this method requires intensive tuning of threshold values [16]. Another study [17] proposes an algorithm that can identify surgical actions more accurately by addressing the limitations of recognizing actions with short duration and non-homogeneous flows. One of the key improvements of the algorithm is the inclusion of semantic visual features, which compensate the kinematic variability. These features are captured from video streams and provide additional information that enhances the classification performance.

The process of surgical task segmentation and action recognition can alternatively be modeled as a sequential decision-making process, in which an agent chooses a segment size and corresponding action label based on a specific policy. In [18], the tree search component considers information about future frames which allows to refine decisions in a reinforcement learning framework for surgical robotic applications.

The integration of kinematic signals with visual features extracted using deep learning models demonstrated improved performance. Nonetheless, the search for optimal segmentation criteria and the need for intensive tuning of threshold values remain a challenge. Further research is necessary to develop more robust and automated methods for surgical task segmentation.

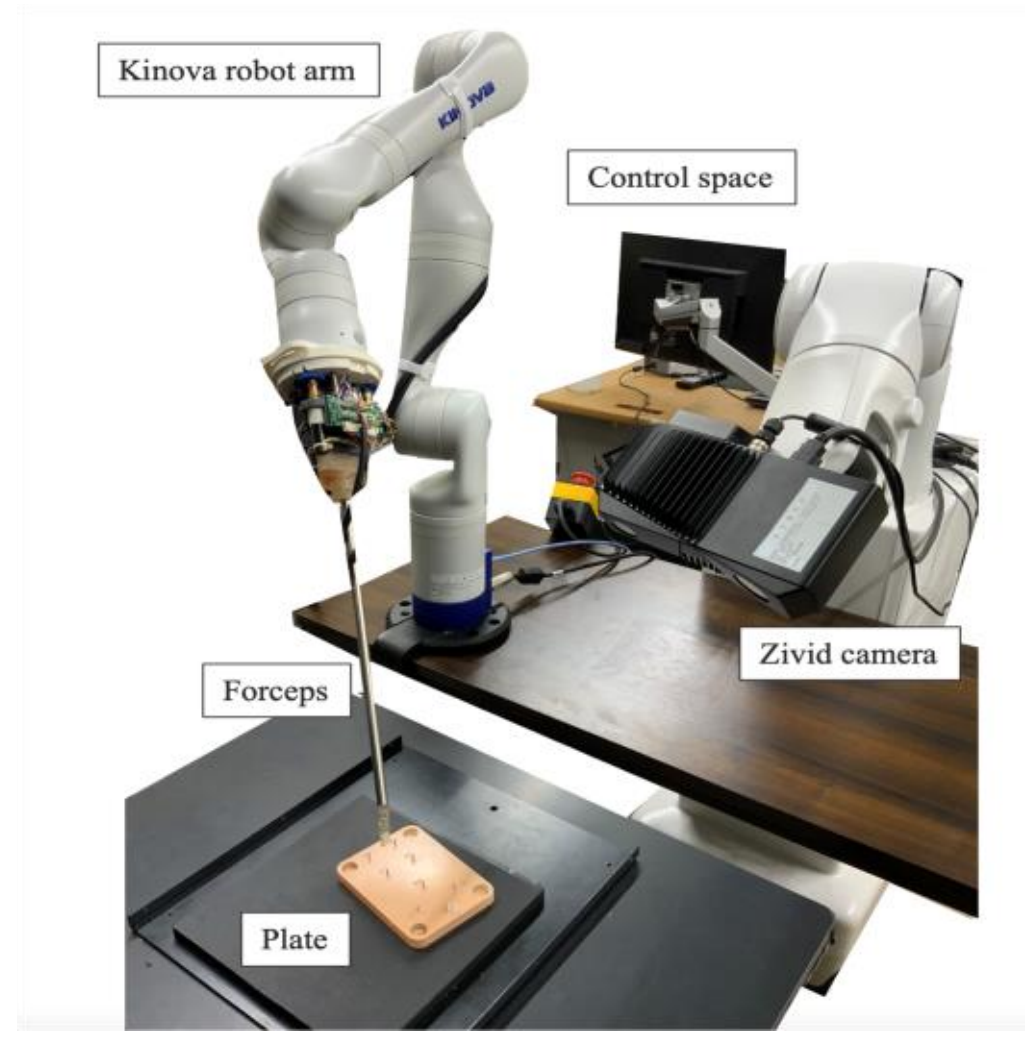

Fig. 1. Robotic setup

## III. Methodology

Our objective is to develop a system that can autonomously assist surgical procedures by recognizing critical transitions in the task. To achieve this, we propose a method that utilizes both visual and kinematic data to identify state transitions in surgical demonstrations. Our approach employs hierarchical clustering to automatically detect key events and changes in the surgical workflow. Once a transition is identified, a predefined robotic surgical tool pose is activated to assist the human surgeon with object manipulation during a pick-and-place surgical task.

### A. *Robotic Setup*

Our robotic setup is shown in Figure 1. It comprises a 7-DOF manipulator (Gen3, Kinova) with a 3-DOF robotic surgical tool attached (OpenRST [19]). Visual information was acquired using a camera (Zivid One +) that provides the endoscope view. The manipulator is teleoperated through a haptic interface (Touch, 3D Systems) without force feedback enabled. The testbed comprises a silicon tube (5 mm diameter) and acrylic base with pegs for tube insertions. The robot controller and transition segmentation algorithm are implemented on a 3.0 GHz Core i9 computer with a GPU (Geforce RTX 3090) running Linux (Ubuntu 20.04, Canonical) and the Robot Operating System (ROS) framework on top of it. The robot control follows an RCM constrained motion control [20] to resemble a surgical operation.

### B. *Dataset*

We select a pick-and-place task that is commonly used in surgical training [21]. Each demonstration starts with the surgical tool in the center of the camera view; the user first grasps the silicon tube located in one of the left pegs and placed it in one of the right pegs, then the surgical tool is returned to the center of the camera view, and finally the user grasps again the silicon tube to return it to the original peg in the left side. The complete sequence can be observed in Figure 2. We collect a set of 14 teleoperated robotic demonstrations $D = \{d_i\}$ from this surgical task. Each demonstrations of a task $d$ is a discrete-time sequence of $T$ state vectors with feature, which is a concatenation of kinematic features $k$ from the proprioceptive robot information and visual features $v$ from the camera.

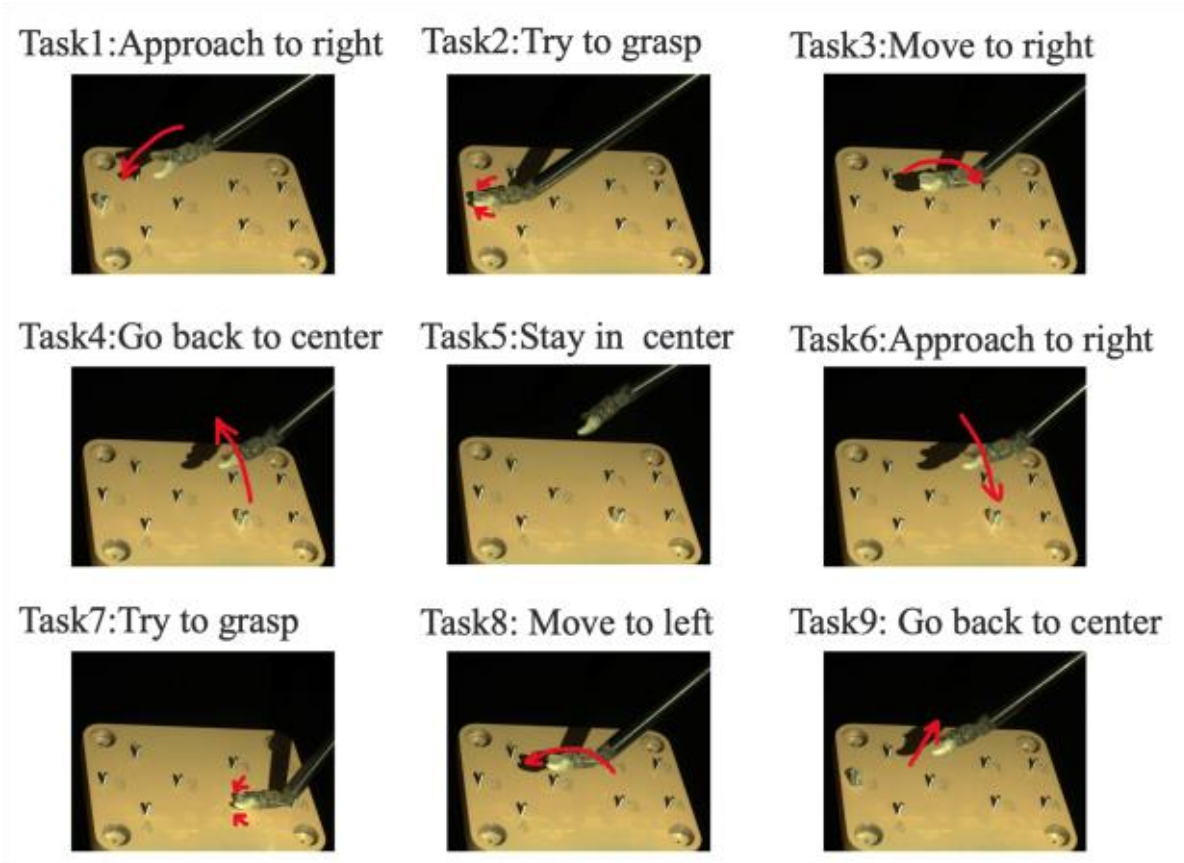

Fig. 2. Sequence of tasks identified in a pick-and-place task.

### C. Visual and Kinematic Features

**Visual Features**: The appropriate choice of visual features has an impact on the segmentation clustering. Here, each T state vector comprises the domain-independent visual features v from the video frames resized to 224x224 pixels. The features were generated by fine-tuning a pretrained Convolutional Neural Networks (CNNs) ResNet18 [22] on the dataset ImageNet-1K [23] to allow extracting useful spatial information for the state representation for the clustering.

**Kinematic Features**: For each $T$ state vector, the robotic tool tip position, linear velocity, angular velocity, tool orientation and gripper angle are utilized as kinematic features $k$.

### D. AutoEncoder algorithm

An AutoEncoder model is tuned with our dataset and provides a dimensional reduced visual feature vector with size 128. The input for our encoder is the feature vector with size 512 generated from the pre-trained ResNet model. AutoEncoder was designed with an encoder of three Fully Connected (FC) layers with 256, 256 and 448 nodes. The decoder also comprises 3 FC layers with 448, 256 and 256. This model utilizes RELU activation functions and RMSprop optimizer. Hyperparameter tuning was performed using Bayesian algorithms from Optuna API [24].

### E. Transition Annotation

Transitions are defined as switching modes of linear dynamical systems. For a demonstration $d_i$ , lets denote transition states $\tau$ composed of the kinematic features and visual features $\tau(t) = (k(t), v(t))$. For our surgical training task, pick-and-place, we identify 9 segments with a total of 8 transition states, as shown in Figure 3. We assumed that the demonstrations are consistent; therefore, the transitions have a partial order and exist nonempty sequence of common transitions.

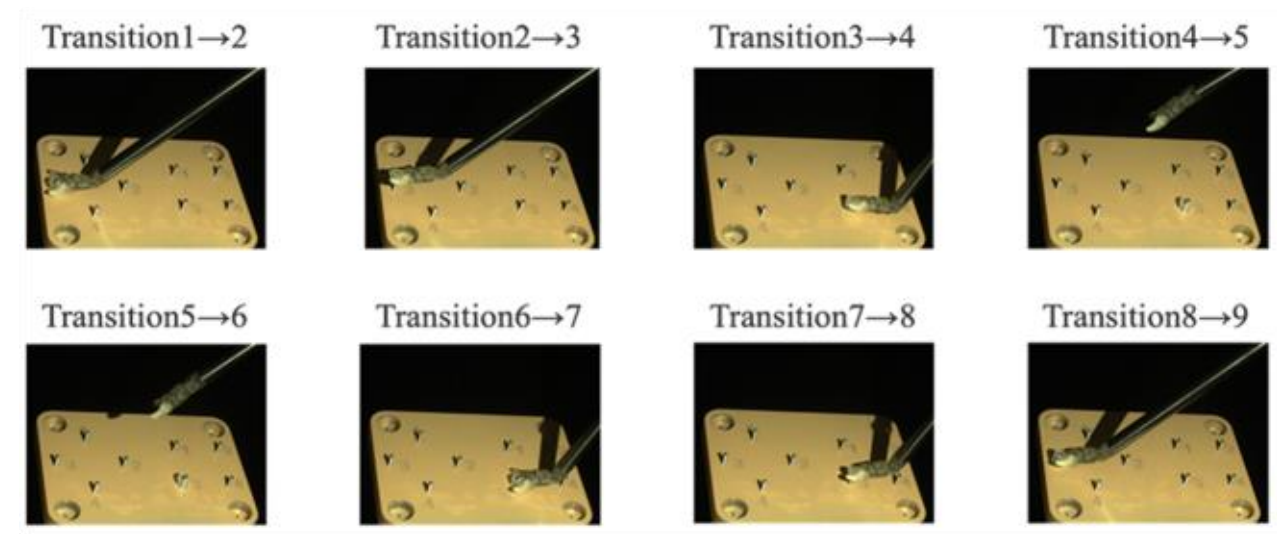

Fig. 3. Transitions identified for a pick-and-place task.

### F. Hierarchical Clustering Algorithm

This method is a form of top-down divisive hierarchical clustering. Given a set of states, a Gaussian Mixture Model (GMM)-based clustering algorithm groups transitions based on first a visual features and then kinematics conditions. The hierarchical clustering generates a parent cluster with varying sizes of sub-clusters. The first visual clustering output is obtained as:

$$C_v = \{(\mu_1, \Sigma_1), (\mu_2, \Sigma_2), \dots, (\mu_l, \Sigma_l)\} \quad (1)$$

Where $\mu_i, \Sigma_i$ represent the mean and covariance matrix of the corresponding Gaussian distribution. For each visual cluster, a subsequent kinematic clustering is obtained as

$$C_i^k = \{(\mu_{i1}, \Sigma_{i1}), (\mu_{i2}, \Sigma_{i2}), \dots, (\mu_{il}, \Sigma_{il})\} \quad (2)$$

Figure 4 shows the proposed scheme of the hierarchical clustering process. The GMM parameters are optimized based on the silhouette score [25], a measure of cluster tightness. For each sub-cluster a transition label is assigned.

### G. Online Autonomous Robotic Assistance

Robotic assistance is defined as virtual fixtures that keep the tool pose in a predetermined orientation to facilitate the grasping and release of the silicon tube. Given a predicted transition, a specific tool orientation is activated that supersedes the current user interface command.

## IV. Transition Segmentation Evaluation

The transition segmentation is validated with a subset of five demonstrations. Examples of two of the demonstrations are shown in Figure 5. Each colored segment corresponds to a different task initialized from a segmented transition. Manual annotations are used as ground truth to compare the predicted transitions. It can be seen that most of the corresponding transitions are correctly identified. Table I summarizes the accuracy performance for the training and testing datasets.

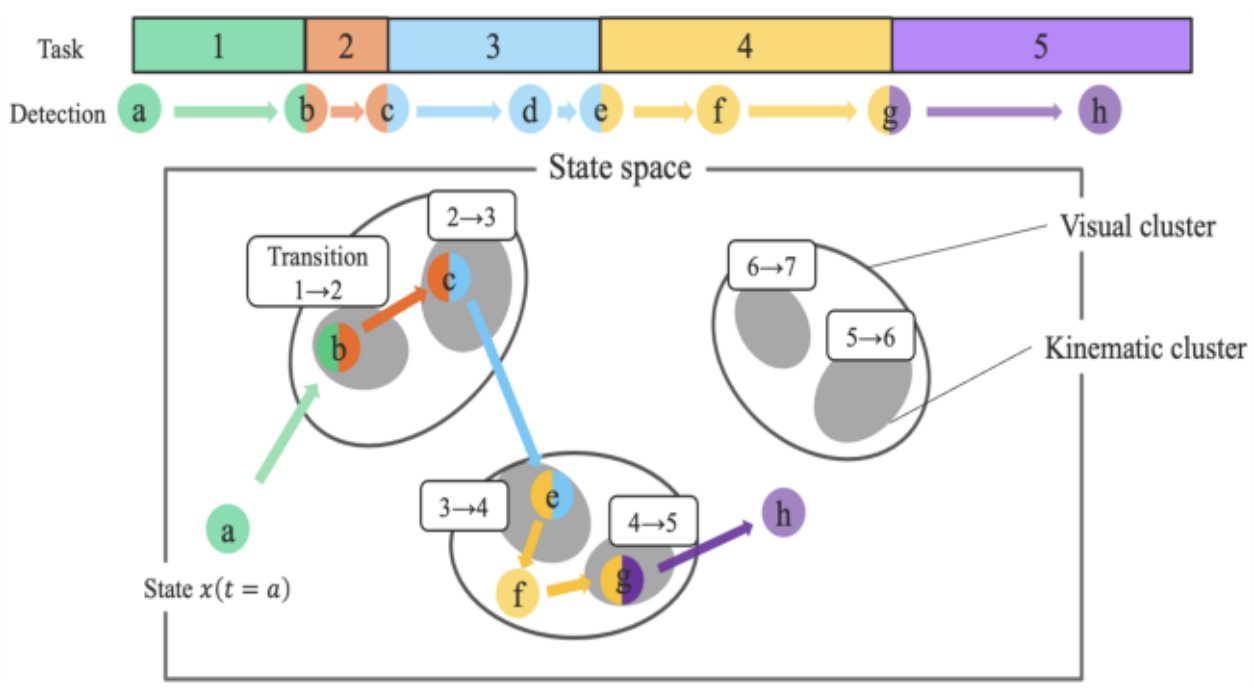

Fig. 4. Hierarchical clustering with a first clustering on visual features, and a second clustering on the kinematic features.

TABLE I. ACCURACY PERFORMANCE

| | Mean | SD |
|---|---|---|
| **Training** | 0.885 | 0.025 |
| **Testing** | 0.853 | 0.016 |

For robot control, online fast predictions are required. We evaluated the time performance required for transition segmentations. Table II describes the time consumed in each module with a total time of 45 ms, sufficient to provide a seamless robotic assistance.

TABLE II. ONLINE SEGMENTATION TIME PERFORMANCE

| Module | Time (ms) |
|---|---|
| Feature extraction | 44.5 |
| Kinematic standardization | 0.2 |
| GMM prediction | 0.5 |
| Robot communication | 0.2 |
| **Total** | **45.5** |

## V. ROBOT-ASSISTANCE EVALUATION

We compared the performance of the task without robot assistance, and with autonomous robot tool pose assistance provided according to the transition detected. Four subjects between the ages of 20 and 25 who had no previous surgical training participated in these experiments. All subjects gave their informed consent for inclusion before they participated in the study. The study was approved by the Ethical Research Committee of Nagoya University.

Each subject was given a 3 trial with and without robot assistance to become familiar with robot operation. The experiment consisted of 3 additional trials with and without assistance. The task completion time is registered as shown in Table III. When robot-assistance is activated, the completion time on average is reduced by about 13%, as shown in Figure 6.

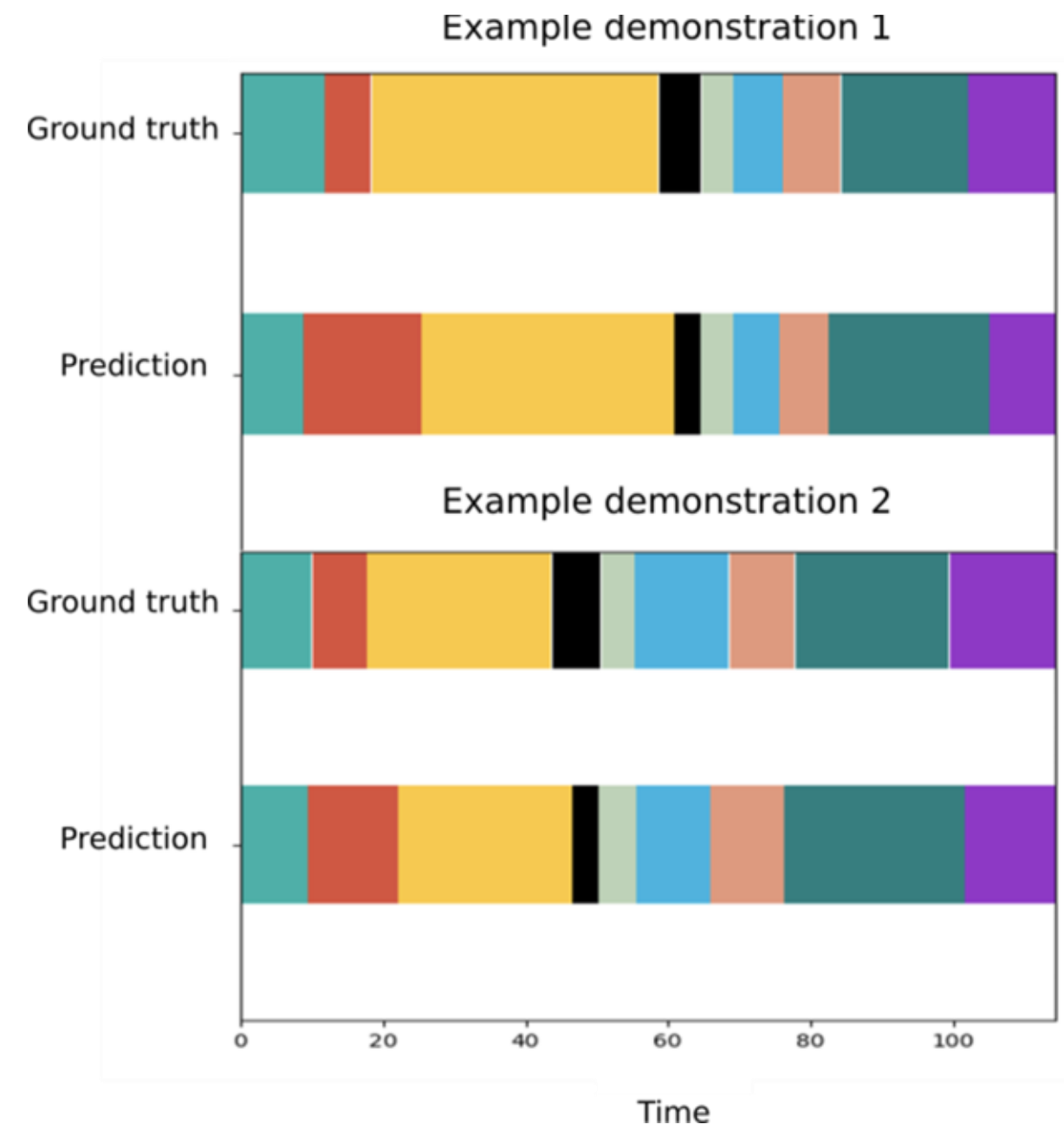

Fig. 5. Examples of task segmentation by transition recognition, with the upper colored bar representing the annotated ground truth and the lower bar the predicted task segmentation.

TABLE III. AVERAGE TASK COMPLETION TIME (S)

| Subject | No assistance | Robot-assisted |
|---|---|---|
| A | 57.9 | 49.6 |
| B | 65.5 | 48.8 |
| C | 65.0 | 57.8 |
| D | 65.8 | 58.8 |
| **All** | **62.5** | **53.9** |

A qualitative evaluation was also carried out based on the NASA-TLX questionnaire to assess cognitive workload when performing a teleoperated pick-and-place task. The weighted workload (WWL) scores are shown in Table IV, with higher values denoting higher mental effort. The robot-assisted mode provides a reduction of about 27% in the cognitive workload.

TABLE IV. WWL ON EACH SUBJECT

| Subject | No assistance | Robot-assisted |
|---|---|---|
| A | 54.2 | 30.5 |
| B | 77.4 | 74.0 |
| C | 41.2 | 24.8 |
| D | 37.9 | 23.8 |
| **All** | **52.6** | **38.3** |

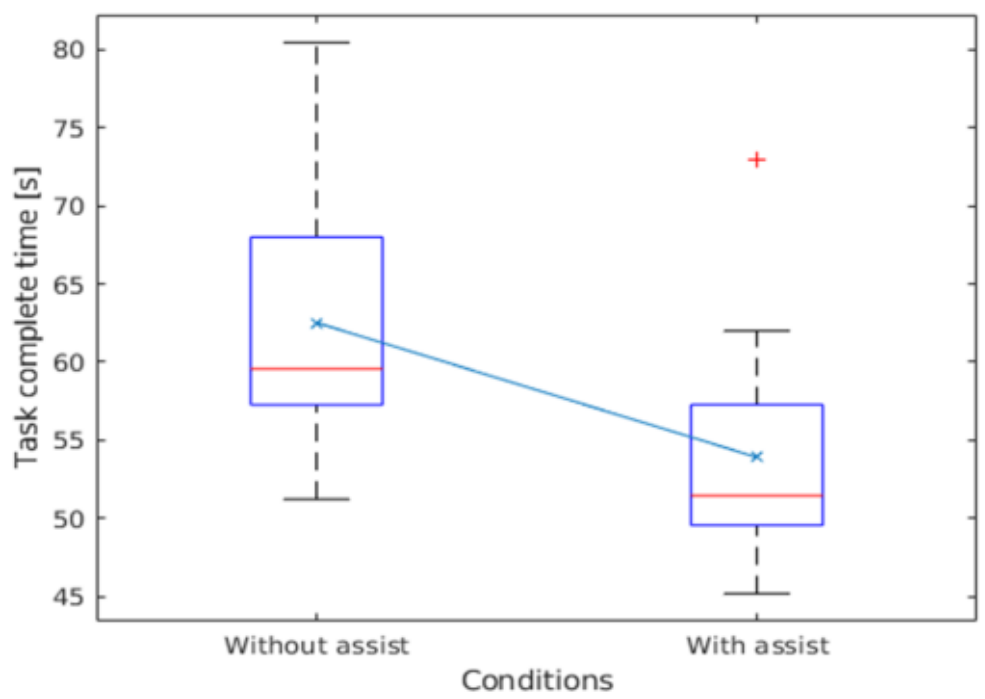

Fig. 6. Task completion time. The use of robot-assistance showed an improvement of about 13%.

## VI. Conclusion

The aim of this study was to investigate the use of online transition state clustering for autonomous robot assistance in a common surgical training task. Since surgical procedures often consist of a series of well-defined tasks, understanding the sequence of tasks is crucial for surgical robotic systems to provide adequate support to surgeons. To achieve this, we proposed a hierarchical approach for clustering visual and kinematic features to recognize task transitions. By monitoring task transitions, the robot can activate predefined assistance, such as tool pose guidance. In experiments, we found that recognizing transitions for robot-assisted activation reduced task completion time and the cognitive workload on the user. Moving forward, we aim to generalize this methodology to more complex surgical tasks, such as suturing and tissue manipulation.

## Acknowledgment

This work was supported in part by the Japan Science and Technology Agency (JST) CREST including AIP Challenge Program under Grant JPMJCR20D5, and in part by the Japan Society for the Promotion of Science (JSPS) Grants-in-Aid for Scientific Research (KAKENHI) under Grant 22K14221.